\documentclass[sigconf]{acmart}

\usepackage{dsfont}
\usepackage{multirow}
\AtBeginDocument{%
  \providecommand\BibTeX{{%
    \normalfont B\kern-0.5em{\scshape i\kern-0.25em b}\kern-0.8em\TeX}}}

\begin{document}

\pagestyle{fancy}
\fancyhf{}
\fancyfoot[C]{\thepage}
%%
%% The "title" command has an optional parameter,
%% allowing the author to define a "short title" to be used in page headers.
\title{ASTRA: An Action Spotting TRAnsformer for Soccer Videos}

%%
%% The "author" command and its associated commands are used to define
%% the authors and their affiliations.
%% Of note is the shared affiliation of the first two authors, and the
%% "authornote" and "authornotemark" commands
%% used to denote shared contribution to the research.
\author{Artur Xarles}
\email{arturxe@gmail.com}
\affiliation{%
  \institution{Universitat de Barcelona \& Computer Vision Center}
  \city{Barcelona}
  \country{Spain}
}

\author{Sergio Escalera}
\email{sescalera@ub.edu}
\affiliation{%
  \institution{Universitat de Barcelona \& Computer Vision Center \& Aalborg University}
  \city{Barcelona}
  \country{Spain}
}

\author{Thomas B. Moeslund}
\email{tbm@create.aau.dk}
\affiliation{%
  \institution{Aalborg University}
  \city{Aalborg}
  \country{Denmark}
}

\author{Albert Clapés}
\email{aclapes@ub.edu}
\affiliation{%
  \institution{Universitat de Barcelona \& Computer Vision Center}
  \city{Barcelona}
  \country{Spain}
}

%%
%% The abstract is a short summary of the work to be presented in the
%% article.
\begin{abstract}
  In this paper, we introduce \textbf{ASTRA}, a Transformer-based model designed for the task of Action Spotting in soccer matches. ASTRA addresses several challenges inherent in the task and dataset, including the requirement for precise action localization, the presence of a long-tail data distribution, non-visibility in certain actions, and inherent label noise. To do so, ASTRA incorporates (a) a Transformer encoder-decoder architecture to achieve the desired output temporal resolution and to produce precise predictions, (b) a balanced mixup strategy to handle the long-tail distribution of the data, (c) an uncertainty-aware displacement head to capture the label variability, and (d) input audio signal to enhance detection of non-visible actions. Results demonstrate the effectiveness of ASTRA, achieving a tight Average-mAP of 66.82 on the test set. Moreover, in the SoccerNet 2023 Action Spotting challenge, we secure the 3rd position with an Average-mAP of 70.21 on the challenge set. 
\end{abstract}

\maketitle

%%
%% Keywords. The author(s) should pick words that accurately describe
%% the work being presented. Separate the keywords with commas.

%%
%% This command processes the author and affiliation and title
%% information and builds the first part of the formatted document.
%

\section{Introduction}

The field of automatic video analysis has significantly impacted the world of sports in recent years. Various computer vision tasks, such as object detection, tracking, and action localization, have found extensive applications within the sports domain. These applications go beyond analyzing player behavior through detection and tracking, encompassing functionalities like automated data collection or video summarization by identifying crucial actions throughout the footage. It is worth noting that this field has witnessed the emergence of numerous tasks and applications, as extensively reviewed by Thomas et al. \cite{thomas2017computer} and Naik et al. \cite{naik2022comprehensive}.\\

This paper specifically focuses on the task of Action Spotting, which involves the temporal localization of multiple actions within untrimmed videos. It shares a close relationship with the well-known task of Temporal Action Localization, differing only in the use of a single keyframe to identify each action. While several sports datasets are available to address this task, covering domains such as tennis~\cite{zhang2021vid2player}, diving~\cite{xu2022finediving}, figure skating~\cite{hong2021video}, and gymnastics~\cite{shao2020finegym}, our primary focus is on soccer. Therefore, to tackle this task, we leverage the SoccerNet-v2 dataset~\cite{soccernet2}, the largest annotated video sports dataset up to date, comprising 550 soccer matches and encompassing 17 distinct actions.\\

To address the task, we propose \textbf{ASTRA} (Action Spotting TRAnsformer), building upon the problem design defined in Soares et al. \cite{yahoo}. This design involves producing time-point detections, each consisting of both a class probability and a temporal displacement over a predefined anchor. ASTRA employs a Transformer encoder-decoder architecture, similar to the one used in DETR \cite{detr}. This allows the model to produce outputs with the desired temporal resolution, regardless of its input temporal resolution. Upon analyzing the dataset, we identify three main challenges: a long-tail distribution of the data, where some actions occur infrequently; the non-visibility of certain actions due to replays or camera angles; and noisy labels resulting from the subjective judgment of annotators in determining temporal locations. To address these challenges, we incorporate different techniques into our model. Firstly, we employ a balanced mixup approach to account for the long-tail distribution of the data. Additionally, we integrate audio signals alongside visual signals to improve the detection of non-visible actions. Furthermore, we introduce an uncertainty-aware displacement head that models label uncertainty using a Gaussian distribution. These techniques enhance performance, with ASTRA achieving an Average-mAP of 66.82 on the test split. We further evaluate our model in the SoccerNet 2023 Action Spotting challenge, consisting of 50 matches with hidden ground-truth, where we achieve the 3rd position with a tight A-mAP of 70.21.\\

The remaining sections of the paper are organized as follows: Section 2 provides a comprehensive review of related work on the task of action spotting. Section 3 introduces ASTRA and outlines its components. Section 4 conducts ablation studies on different aspects of the model, and compares our best solution against state-of-the-art works. Finally, Section 5 concludes the paper, summarizing our key findings and conclusions derived from this research.
\vspace{-0.1cm}

\section{Related work}

{\large \textbf{Temporal Action Localization \& Action Spotting.}} Action recognition has undergone significant advancements in recent years, playing a crucial role in video understanding. Initially, methods focused on classifying short trimmed videos~\cite{kinetics, youtube8m, somethingsomething}. However, with the progress in computer vision, more challenging tasks have emerged. Two prominent tasks in this domain are Temporal Action Localization (TAL) and Action Spotting (AS), which share the objective of temporally locating multiple actions within untrimmed videos. While TAL represents actions as temporal intervals through the annotation of \textit{begin} and \textit{end} frames, AS represents actions with a \textit{single keyframe}. This distinction offers an advantage for AS in terms of annotation cost, as it requires only one frame per action. Moreover, AS is especially well-suited for capturing actions that are instantaneous or have uncertain start and end times, where a single timestamp can effectively represent them. A concrete example is demonstrated in the SoccerNet-v2 dataset~\cite{soccernet2}, where actions like goals or fouls are typically identifiable at specific temporal points. \\

Given the inherent similarities between TAL and AS, the methods developed for these tasks often share common components, with their main differences lying in the prediction head. These methods can generally be categorized into two groups: two-stage methods~\cite{heilbron2016fast, escorcia2016daps, buch2017sst, zhou2021feature, qing2021temporal, xu2020g} and one-stage methods~\cite{lin2021learning, e2espot, yahoo, calf, actionformer, tridet, tadtr}. In two-stage methods, proposals are first generated and subsequently classified to determine if they correspond to actions or background. These methods tend to be more complex and do not allow for end-to-end training. In contrast, one-stage models directly localize and classify actions in a single step, eliminating the need for proposal generation. These models offer simplicity and often achieve state-of-the-art performance on TAL and AS tasks. \\

Early one-stage models in temporal action localization utilized anchor windows sampled from sliding windows~\cite{buch2019end, lin2017single}. For instance, Lin et al.~\cite{lin2017single} employed a set of anchor windows that were classified into different categories and refined using location offsets and overlap scores. Later, Yang et al.~\cite{yang2020revisiting} introduced an anchor-free approach that relied on temporal points instead of anchor windows for action localizations. Their work showed the benefits of both anchor-free and anchor-based approaches. Current state-of-the-art methods in temporal action localization are predominantly anchor-free. In particular, ActionFormer~\cite{actionformer} and TriDet~\cite{tridet} have achieved remarkable performance in this field. They classify each moment as either background or one of the possible actions. ActionFormer utilizes a transformer encoder architecture with downsizing operations, while TriDet incorporates a Scalable-Granularity Perception (SGP) layer based on CNNs. The SGP layer replaces the self-attention mechanism of ActionFormer to improve both model performance and efficiency. These approaches also utilize temporal regression to refine predictions and obtain more precise results. Another method, TadTR~\cite{tadtr}, draws inspiration from the DETR model~\cite{detr} for object detection. TadTR constructs a transformer encoder-decoder architecture with learnable queries representing detection candidates. During training, a bipartite matching problem pairs those candidates with ground-truth actions.\\

Similar techniques have also demonstrated state-of-the-art performance in the task of action spotting on SoccerNet, as further discussed in Giancola et al.~\cite{giancola2022soccernet}. For instance, E2E-Spot~\cite{e2espot} proposes a 2D CNN backbone with Gate Shift Modules~\cite{sudhakaran2020gate}, which incorporate temporal context and produce per-frame predictions using a Gated Recurrent Unit~\cite{cho2014learning} layer. This model operates directly on the raw video frames, providing increased flexibility compared to using pre-extracted features. However, that introduces additional complexity and computational cost during training. Soares et al.~\cite{yahoo} achieve SOTA results by defining a set of dense anchors (i.e. one anchor per input token), similar to ActionFormer or TriDet, to represent temporal positions. These anchors are then classified into different action classes and temporally refined. The model uses pre-extracted features obtained from various pre-trained video backbones and utilizes a U-Net-like architecture for the model's trunk.\\

Our approach takes inspiration from the Transformer encoder-decoder architecture of TadTR and DETR. We employ a similar architecture, with learnable queries in the decoder, where each generated query represents a specific temporal position, akin to the dense anchors proposed in Soares et al.'s work. Furthermore, we also leverage a set of strong pre-extracted features to train our model, providing a solid foundation for accurate action localization, and avoiding the added complexity when using raw frames. \\

\noindent{\large\textbf{Uncertainty estimation.}} Uncertainty estimation techniques have demonstrated their potential to enhance the performance of regression models by providing reliability estimates and accounting for potential errors in predicted values~\cite{AQAuncertainty, xie2020boundary, chen2020refinement}. This becomes particularly valuable when dealing with inherently uncertain ground-truth data, characterized by measurement errors, noise, or label ambiguity. For instance, Tang et al.~\cite{AQAuncertainty} approached Action Quality Assessment (AQA) task by modelling the quality score as a Gaussian distribution, maximizing the log-likelihood function to estimate both mean and variance. Similarly, Xie et al.~\cite{xie2020boundary} and Chen et al.~\cite{chen2020refinement} also employed Gaussian distributions for temporal regression in TAL. However, they used the Kullback-Leibler divergence to fit their models. In our work, we adopt a similar Gaussian distribution for modelling temporal displacements, and like in~\cite{AQAuncertainty}, we maximize the log-likelihood function for fitting purposes. \\

\noindent{\large \textbf{Multimodal approaches.}} In addition to the visual modality, certain methods for action classification, TAL or AS incorporate additional modalities. These modalities can include optical flow~\cite{wang2016temporal, zhou2018temporal, lin2019tsm} or audio~\cite{vanderplaetse2020improved, shaikh2022maivar, pieropan2014audio, kazakos2019epic} among others, and they differ in how they fuse these modalities. Specifically, for AS in SoccerNet, an approach that combines different modalities is the one in Vanderplaetse and Dupont~\cite{vanderplaetse2020improved}. They extract features from the log-mel spectrogram of the audio using a VGG-inspired model and explore various fusion techniques. In our work, we adopt a similar approach, leveraging a VGG-inspired model to extract audio features from the log-mel spectrogram. However, we further fine-tune the backbone model during training. We perform an early fusion of different features, merging them at the input of the Transformer encoder. 

\begin{figure*}[ht]
	\centering
	\includegraphics[width=0.9\linewidth]{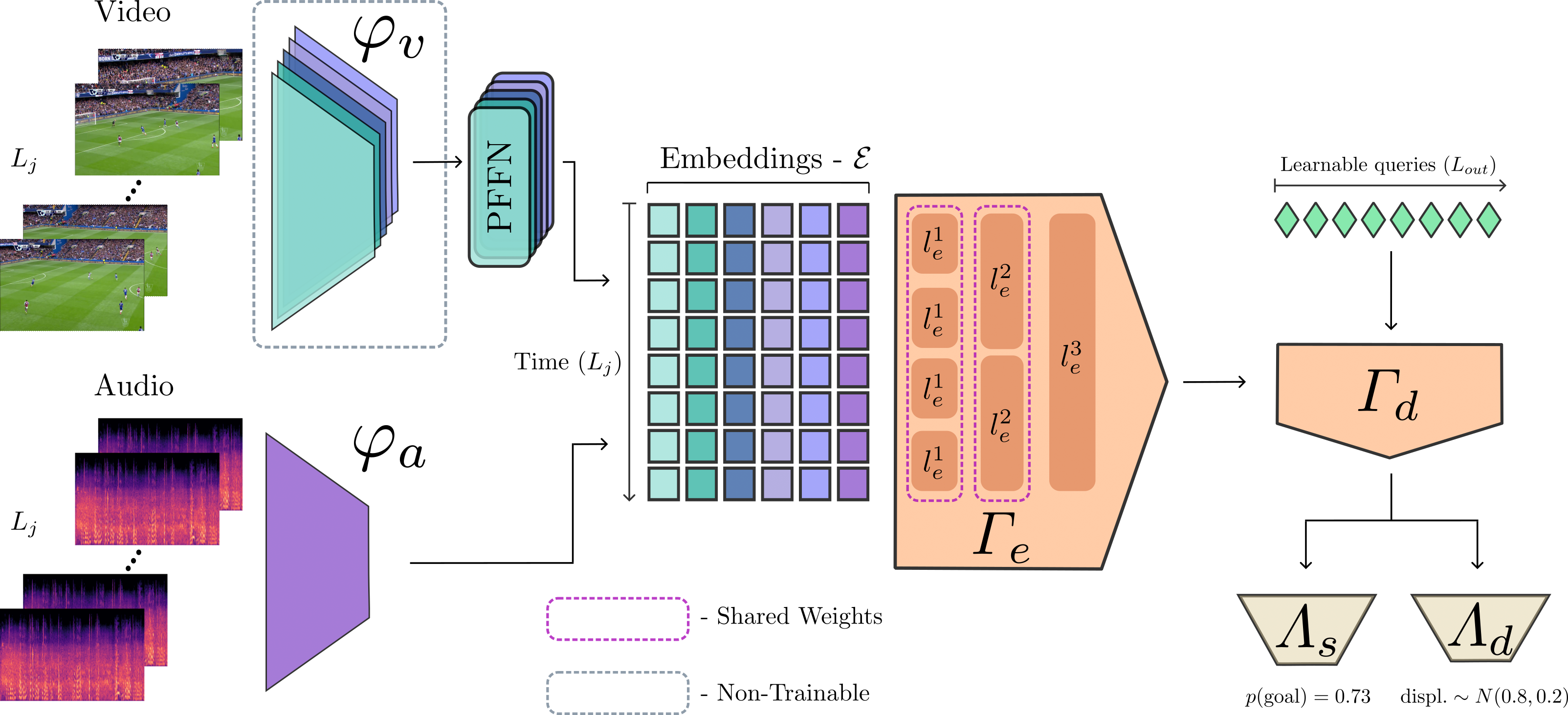}
	%   \caption{Base model for action spotting. Multimodal transformer with audio and video features and intermediate supervision.}
	\caption{ASTRA (Action Spotting TRAnsformer) architecture: Visual and audio embeddings from different backbones ($\varphi_v$ and $\varphi_a$) are combined and processed through a Transformer encoder-decoder ($\varGamma_e$ and $\varGamma_d$). The resulting embeddings are then utilized by a classification head ($\varLambda_s$) for temporal location classification, and a displacement head ($\varLambda_d$) to further refine predictions.}
	\label{fig:ASTRA}
\end{figure*}

\section{Methods}

{\large\textbf{Problem definition.}} Action spotting involves the identification and precise localization of actions within an untrimmed video $X$. Given the video input or a representation of it, the objective is to identify and locate all the actions occurring in the video, represented as $A = \{a_{1}, \dots, a_{N}\}$. The number of actions, denoted as $N$, may vary across different videos. Each action instance $a_{i}$ comprises an action class $c_{i}$ and its corresponding temporal position $t_{i}$, forming a pair $a_{i} = (c_{i}, t_{i})$. Here, $c_{i} \in \{1, \dots, C\}$ represents the action class index, with $C$ being the total number of distinct action classes.\\

\noindent {\large\textbf{Method overview.}} 
Our solution, ASTRA, leverages embeddings $\mathcal{E}$ from multiple modalities to achieve its goals. Specifically, ASTRA is built upon $|\mathcal{E}|-1$ pre-computed visual embeddings, complemented by an additional audio embedding derived from the log-mel spectrogram of the audio using a VGG-inspired backbone. The network responsible for generating this audio embedding is jointly trained with the ASTRA model. The embeddings are input to the model in clips spanning a duration of $T$ seconds. These features from each backbone are processed in parallel streams, where Point-wise Feed-Forward Networks (PFFN) project them to a common dimension $d$. The projected embeddings are then combined in the subsequent Transformer encoder-decoder module, with learnable queries in the decoder. Inspired by the architecture proposed in DETR, this module enables ASTRA to handle different input and output temporal dimensions ($L_{in}$ and $L_{out}$, respectively) and facilitates a straightforward fusion of multiple embeddings. To enhance ASTRA's ability to capture fine-grained details, we introduce a temporally hierarchical architecture for the Transformer encoder. This architecture enables the encoder to attend to more local information in the initial layers and reduces the computational cost. Finally, ASTRA employs two prediction heads to generate classification and displacement predictions for the anchors introduced by Soares et al.~\cite{yahoo}. These anchors correspond to specific temporal positions and class actions, as described in their work. Additionally, we adopt their suggestion of employing a radius for both classification and displacement ($r_{c}$ and $r_{d}$, respectively) to define the temporal range around a ground-truth action within which it can be detected.

Furthermore, to account for label uncertainty, ASTRA adapts the prediction head responsible for displacement by modeling them as Gaussian distributions instead of deterministic temporal positions. This allows ASTRA to capture temporal location uncertainty and provide a more comprehensive representation of the actions. Additionally, ASTRA incorporates a balanced mixup technique to improve model generalization and accommodate the long-tail distribution of the data.

We illustrate the ASTRA architecture in Figure \ref{fig:ASTRA}, and further details are discussed in the subsequent sections. 

\subsection{Input embeddings}

As previously mentioned, ASTRA is built upon $|\mathcal{E}|$ embeddings proceeding from multiple modalities, $|\mathcal{E}|-1$ pre-extracted visual embeddings and an additional audio embedding. Let $E_{j} \in \mathbb{R}^{L_{j}\times d_{j}}$ denote the sequence of features associated to the $j$-th embedding, where $j\in\{1, ..., |\mathcal{E}|\}$. Here, $L_{j}$ represents the temporal dimension (i.e. $L_{in}$ for each embedding), and $d_{j}$ represents the feature dimension specific to that embedding. It is important to note that different embeddings may have varying temporal or feature dimensions.\\

\noindent {\large\textbf{Visual embeddings.}} The $|\mathcal{E}|-1$ visual embeddings used as inputs to our model are obtained from the Baidu Research repository.\footnote{\url{https://github.com/baidu-research/vidpress-sports}} They are extracted using distinct backbones ($\varphi_{v}$) fine-tuned on the SoccerNet dataset for action classification. With a receptive field of 5 seconds, each embedding is computed with a stride of 1. To ensure a consistent feature dimension $d$ across all embeddings, PFFNs are employed. These PFFNs project the embeddings through two linear layers with ReLU activation, applying dropout with probability $p$. \\

\noindent {\large\textbf{Audio embedding.}} For the additional audio embedding, we employ a VGG-inspired backbone~\cite{vggish} ($\varphi_{a}$), which is pre-trained on the AudioSet dataset~\cite{audioset}. The backbone takes the log-mel spectrogram of the audio as input and is fine-tuned during the training of the ASTRA model. We further replace the last linear layer of the backbone to produce the desired feature dimension of $d$. In line with common practice, we feed the backbone with log-mel spectrogram segments, each spanning a duration of $T_{a}$ seconds. Consequently, we obtain the audio embedding $E_{a}$ with $L_{a} = \lfloor T / T_{a}\rfloor$ and $d_{a} = d$.

\subsection{Transformer encoder-decoder}

After aligning the feature dimension of all embeddings in $\mathcal{E}$, they are passed into the Transformer encoder-decoder module. Prior to that, a learnable encoding specifying temporal position and backbone source is added. Then, the enriched tokens (i.e., feature vectors corresponding to specific embeddings and temporal positions) undergo a Hierarchical Transformer encoder, where they progressively interact with tokens that are further apart in terms of temporal distance. This hierarchical structure enables the model to attend to fine-grained local details in the early layers while gradually incorporating broader context in the subsequent layers. In the Transformer decoder, a set of $L_{out}$ learnable queries, representing temporal positions, is introduced. These queries evolve and capture relevant information from the Transformer encoder output tokens during the self-attention and cross-attention mechanisms in the decoder. \\

\noindent {\large\textbf{Hierarchical Transformer encoder ($\varGamma_e$).}} The Hierarchical Transformer encoder is composed of $n_{e}$ vanilla Transformer encoder layers. Each layer applies the standard multi-head self-attention with $h$ heads, followed by a two-layered PFFN with a widening factor of $\alpha$, dropout, layer normalization, and residual connections. To incorporate the temporal hierarchy, in each layer $l_{e}^{i}$, where $i\in \{1, \dots, n_{e}\}$, the input clip of $T$ seconds is divided into $2^{n_{e}-i}$ segments. Tokens within the same temporal segment are processed together within the layer. Importantly, all segments within a layer share the same transformer encoder layer, ensuring weight sharing and parameter efficiency.\\

\noindent {\large\textbf{Transformer decoder ($\varGamma_d$).}} The transformer decoder is composed of $n_{d}$ vanilla Transformer decoder layers. Each layer applies the standard multi-head self-attention and multi-head cross-attention, each with $h$ heads, followed by a two-layered PFFN with a widening factor of $\alpha$, dropout, and residual connections. Unlike the hierarchical structure in the encoder, in the decoder, all tokens within the same clip interact with each other.\\

The Transformer encoder-decoder module in ASTRA provides two main advantages over other methods in TAL or AS:
\begin{enumerate}
    \item Flexible handling of input and output temporal dimensions, $L_{in}$ and $L_{out}$. While the input temporal dimension is typically fixed, ASTRA allows for a different output temporal dimension, providing the ability to customize the temporal resolution. This flexibility is particularly beneficial in our AS task, as highlighted in Section \ref{sec:ablations}.

    \item Seamless integration of multiple embeddings with varying temporal dimensions. Unlike other methods that require embeddings to have the same temporal dimension and concatenate them along the feature dimension, ASTRA can accommodate individual embeddings as separate tokens, allowing for diverse temporal resolutions. 
\end{enumerate}

\subsection{Prediction heads}

\begin{figure}[ht]
	\centering
	\includegraphics[width=1\linewidth]{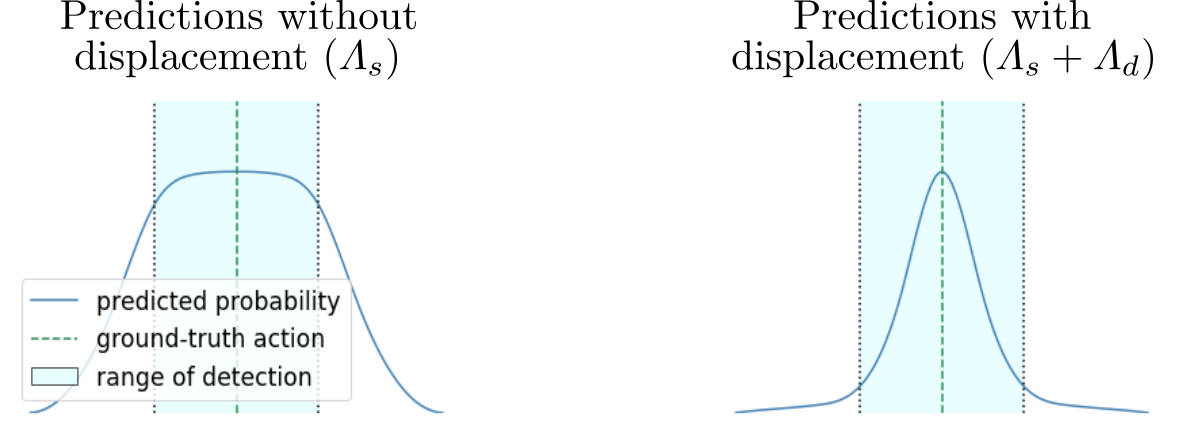}
	%   \caption{Base model for action spotting. Multimodal transformer with audio and video features and intermediate supervision.}
	\caption{Example ground-truth action prediction with and without displacements. The utilization of only the classification head results in predictions spanning the entire range of detection of the ground-truth action (left), whereas incorporating displacements refines the predictions, aligning them closer to the actual temporal position of the ground-truth action (right). }
	\label{fig:ASTRA_heads}
\end{figure}

The evolved queries, representing $L_{out}$ temporal positions uniformly distributed over the $T$ seconds, are input to two prediction heads: the classification head and the uncertainty-aware displacement head. Figure \ref{fig:ASTRA_heads} provides a visual representation of the predictions produced by these prediction heads. \\

\noindent {\large\textbf{Classification head ($\varLambda_s$).}} The classification head consists of two linear layers that project the evolved queries to the desired output feature dimension $C+1$, representing the $C$ different actions plus an additional background class. We incorporate dropout of $p$ and use the ReLU activation function as an intermediate activation. Finally, a sigmoid activation function is applied so the output for each pair of temporal position and action class represents the probability of a ground truth action of that class occurring within the range of detection of the corresponding temporal position.\\

\noindent {\large\textbf{Uncertainty-aware displacement head ($\varLambda_d$).}} The uncertainty-aware displacement head is composed of two linear layers with a dropout rate of $p$, using the ReLU activation function. Two additional linear layers are constructed in parallel, taking the previous output as input. Both layers project the evolved queries to a feature dimension of $C+1$ and apply dropout. The first layer utilizes a linear activation function and outputs the estimated mean displacement for each query and action class. The second layer employs an exponential activation function to generate positive values representing the estimated variance. This allows us to model the displacement as a Gaussian distribution, capturing the uncertainty associated with the displacement estimation for each temporal position and action class. These estimated displacements are then used to refine the predictions produced by the classification head. \\

In summary, the prediction heads produce predictions $\hat{y}_{i}^{t, c} = (\hat{s}_{i}^{t, c}, \hat{d}_i^{t, c}\sim N(\{\hat{\mu}_{i}^{t, c}, \hat{\sigma}_{i}^{t, c}\}))$ for each temporal position $t \in \{1, ..., L_{out}\}$ and action class $c \in \{1, ..., C+1\}$ in a given clip sample $i$. The first element corresponds to the classification score, while the second represents the estimated Gaussian distribution for the displacement, indicated by the predicted mean and variance.

\subsection{Data augmentation}
\label{sec:augmentation}
ASTRA is strengthened by integrating a diverse set of data augmentation techniques applied to the input features. These techniques are designed to improve the model's generalization capability and, for some of them, to also account for the long-tail distribution of the data. These techniques are as follows:  \\

\noindent {\large\textbf{Balanced mixup.}} Similar to traditional mixup~\cite{mixup}, virtual training examples are generated by creating linear combinations of pairs of examples using a parameter $\lambda$ sampled from a beta distribution $\lambda \sim Beta(\alpha_{m}, \beta_{m})$. However, our approach has a distinction. Instead of sampling both samples from the same original distribution, the second element of the pair is sampled from a balanced distribution created using a queue. This queue contains two samples from each action class and is updated at each batch iteration.\\

%\begin{equation}
%\label{eq:mixup}
%\begin{split}
%\tilde{x} = \lambda x_{i} + (1- \lambda) x_{j}\\
%\tilde{y} = \lambda y_{i} + (1-\lambda) y_{j}
%\end{split}
%\end{equation}

\noindent {\large\textbf{Temporal dropout and temporal switch.}} Treating a temporal sequence as the set of tokens corresponding to the same temporal position, we introduce two techniques. Firstly, in temporal dropout, we randomly drop entire temporal sequences with probability $p_{td}$. In the dropped positions, we substitute them with learnable tokens. Secondly, in temporal switch, we randomly swap the positions of consecutive pairs of temporal sequences with probability $p_{ts}$.

\subsection{Training details}

The model is trained using a combination of a classification loss ($\mathcal{L}_{c}$) and a displacement loss ($\mathcal{L}_{d})$. For classification, we employ a binary cross-entropy focal loss for all actions, temporal positions, and data samples, as formulated in Equation \ref{eq:lossC}, where ground-truth labels are denoted as $y_{i}^{t, c} = (s_{i}^{t, c}, d_{i}^{t, c})$. The hyperparameter $\gamma$ adjusts the rate at which easy examples are down-weighted.

\begin{equation}
\label{eq:lossC}
\resizebox{.9\hsize}{!}{$
\begin{split}
\mathcal{L}_{c} = - \frac{1}{N\cdot L_{out} \cdot (C+1)} \Bigl(  \sum_{i=1}^{N} \sum_{t=1}^{L_{out}}\sum_{c=1}^{C+1}  |s_{i}^{t, c} - \hat{s}_{i}^{t, c}|^{\gamma} \cdot \bigl( s_{i}^{t, c} \cdot \ln(\hat{s}_{i}^{t, c}) \hspace{2mm} +  \\  (1-s_{i}^{t, c}) \cdot \ln(1-\hat{s}_{i}^{t, c}) \cdot  \bigl) \Bigl)
\end{split}$}
%\vspace{0.2cm}
\end{equation}

For the displacement loss ($\mathcal{L}_{d}$), it is only applied within the $r_{d}$ seconds radius of ground-truth actions and is based on the negative log-likelihood function of the target Gaussian distribution. We formulate $\mathcal{L}_{d}$, similar to Zhang et al. \cite{AQAuncertainty}, as shown in Equation \ref{eq:lossD}.

\begin{equation}
\label{eq:lossD}
\resizebox{.9\hsize}{!}{$
\begin{split}
\mathcal{L}_{d} = - \frac{1}{N_{dis}} \cdot \sum_{i=1}^{N} \sum_{t=1}^{L_{out}}\sum_{c=1}^{C+1} \mathds{1}_{\{\exists d_{i}^{t, c}\}} \cdot \Bigl(  \frac{\alpha_{L}}{(\hat{\sigma}_{i}^{t, c})^{2}} \cdot |d_{i}^{t, c} - \hat{\mu}_{i}^{t, c}|^{2} \hspace{2mm} + \\ (1 - \alpha_{L}) \cdot \ln [ (\hat{\sigma}_{i}^{t, c})^{2} ] \Bigl)
\end{split}$}
%\vspace{0.2cm}
\end{equation}

In the above equation, $N_{dis}$ is the total number of ground-truth displacements (i.e. inside the range of detection of a ground-truth action). Additionally, $\alpha_{L}$ is a weight that balances the attention paid to uncertain information, with larger values of $\alpha_{L}$ focusing more on the uncertainty, while smaller values tend to result in a more typical single point estimation of the displacement. 

To effectively merge both losses, we introduce a weight $w_{c}$ on the classification loss. This weight ensures that both losses are scaled to the same range of values.

%\begin{equation}
%    \label{eq:lossF}
%    \mathcal{L} = w_{c} \cdot \mathcal{L}_{c} + \mathcal{L}_{d}
%\end{equation}

\subsection{Inference}

At inference time, the data augmentation techniques are disabled. Moreover, the temporal position classifications are refined by incorporating the displacement estimations, represented by the mean of the estimated Gaussian distribution. To reduce the number of candidate actions, Soft Non-Maximum Suppression~\cite{bodla2017soft} is applied with a 1D adaptation as proposed in the work by Soares et al. \cite{yahoo}.

\section{Results}

In this section, we provide an overview of the dataset used in our study, highlighting its key characteristics and challenges. We also discuss the implementation details, the evaluation metric and protocols employed for assessing the proposed models, and present a comprehensive analysis of all ablation experiments conducted. Lastly, we provide a detailed evaluation of our best-performing model, including its performance on the \href{https://www.soccer-net.org/challenges/2023}{2023 SoccerNet challenge}. 

\subsection{Dataset}

\begin{table*}[ht]
	\centering
	\scalebox{0.9}{
		\begin{tabular}{|l|ccccccccc|}\hline
                Action & Ball out of play & Throw-in & Foul & Indirect FK & Clearance & Shot on target & Shot off target & Corner & Substitution \\
                \hline
                Absolute frequency & 31810 & 18918 & 11674 & 10521 & 7896 & 5820 & 5256 & 4836 & 2839 \\
                \hline \hline
                Action & Kick-off & Direct FK & Offside & Yellow Card (YC) & Goal & Penalty & Red Card (RC) & YC -> RC & \\
                \hline
                Absolute frequency & 2566 & 2200 & 2098 & 2047 & 1703 & 173 & 55 & 46 &\\
                \hline
			
		\end{tabular}
		\vspace{-0.5cm}
	}
	\caption{Frequency of occurrence of the different actions in the 500 games with publicly available annotations.}
	\label{tab:nactions}
    \vspace{-0.7cm}
\end{table*}

SoccerNet-v2 is a comprehensive dataset comprising 550 soccer matches from major European competitions. Among these matches, 500 have publicly available annotations with keyframes depicting 17 different actions. Table \ref{tab:nactions} provides a breakdown of these actions and their frequencies in the annotated matches. The remaining 50 matches serve as a hidden ground-truth evaluation set, accessible only to the organizers for assessing the submitted predictions. 

While solving the task of action spotting for this dataset, we encounter three main difficulties:\\

\noindent {\large\textbf{Long-tail data.}} Like many real-world datasets, SoccerNet exhibits a highly unbalanced distribution. As shown in Table \ref{tab:nactions}, certain actions occur much more frequently than others. This imbalance poses a challenge, as a model that disregards the long-tail distribution may perform well on the predominant head classes but struggle to adequately address the less frequent tail classes. Overfitting to these tail classes is also a concern. This issue is further aggravated when considering the task of classifying every temporal position $L_{out}$ and introducing an additional background class. To mitigate this problem, our approach incorporates balanced mixup. This technique forces the model to iterate more times through clips (or mixtures of clips) containing actions, particularly those from the tail end of the distribution. By leveraging this approach, our aim is to enhance the model's ability to handle long-tail actions, while simultaneously improving its generalization capabilities across all classes, as typically observed in mixup approaches. \\

\begin{figure}[ht]
	\centering
	\includegraphics[width=1\linewidth]{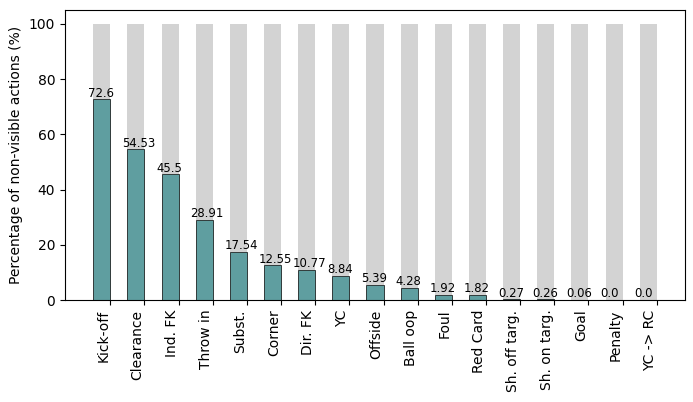}
	%   \caption{Base model for action spotting. Multimodal transformer with audio and video features and intermediate supervision.}
        \vspace{-0.7cm}
	\caption{Percentage of non-visible ground-truth actions for each action class.}
        
	\label{fig:nonvisible}
    \vspace{-0.4cm}
\end{figure}

\noindent {\large\textbf{Non-visible actions.}} In the original videos from SoccerNet, not all annotated actions are directly visible due to replays or camera transitions that may occlude the actions. This is accentuated in some actions as kick-offs, clearances or indirect free-kicks, as illustrated in Figure~\ref{fig:nonvisible}. Consequently, the model needs to rely on contextual information and extrapolation to predict these actions. To address this challenge, we incorporate audio into the model, assuming that the broadcast commentary or the audience reactions can assist in identifying some of the actions that are not visually observable in the videos. \\

\noindent {\large\textbf{Noisy labels.}} Annotating actions can be subjective, leading to varying degrees of clarity in indicating the exact temporal positions of actions. While some actions have distinct temporal indications, others are more complex and rely on the annotator's judgement. This subjectivity introduces noise in the temporal annotations. Furthermore, the presence of non-visible actions exacerbates the annotation challenge, as the temporal selection of these actions relies solely on the annotator's subjective judgment. To address these issues, we introduce uncertainty estimation techniques. We incorporate an uncertainty-aware displacement head that models displacements as Gaussian distributions rather than deterministic values. By capturing the inherent uncertainty in the ground-truth data, we aim to mitigate the impact of noisy labels.

\subsection{Evaluation metric \& Evaluation protocols}

\noindent {\large\textbf{Evaluation metric.}} The performance of the model's spotted actions is assessed using the Average-mAP. This metric quantifies the Area Under the Curve (AUC) of the mean Average Precision (mAP) at different tolerances $\delta$. The mAP is computed by averaging the Average Precision (AP) values across different action classes. The AP summarizes the precision-recall curve into a single value, representing the average precision across all recalls. It can be computed as follows:
	
\begin{eqnarray}
	\label{eq:AP}
	AP = \sum_{s=0}^{S-1} (Recalls(s) - Recalls(s+1)) \cdot Precisions(s)
\end{eqnarray}

Here $S$ denotes the total number of thresholds considered, while $Recalls(s)$ and $Precisions(s)$ refer to the recall and precision, respectively, at threshold $s$. We can further denote Average-AP as the per-class Average Precision averaged across the different tolerances. 

In SoccerNet, two versions of this metric are commonly employed. A loose metric with tolerances ranging from 5 to 60 seconds, and a tight metric with tolerances between 1 and 5 seconds. We present results for both metric versions, but we primarily focus on the tight metric to guide decisions regarding our model's components because it aligns with the SoccerNet 2023 challenge.\\

\noindent {\large\textbf{Evaluation protocols.}} We employ two different evaluation protocols to train and evaluate our models:

\begin{itemize}
    \item \textbf{Protocol 1 (P1):} The dataset of 500 annotated matches is divided into three sets: train, validation, and test. The train split is used for model training, the validation split is utilized to determine the optimal stopping point during training, and the test split is employed to quantitatively compare different models in our ablation experiments, evaluating their performance using the previously introduced metric. Each model is trained five times with different seeds, and the average of these runs is reported. 
    \item \textbf{Protocol 2 (P2):} The model selected based on P1 is trained using the combined data from train, validation and test sets. Subsequently, the trained model is used to generate predictions on the challenge split and obtain the final performance.
\end{itemize}

\subsection{Implementation details}

We employed PyTorch for model implementation, using Adam optimizer with base LR of $5e^{-5}$. Optimization encompassed Learning Rate Warmup via Cosine Decay over 50 epochs, with 3 epochs for initial warmup. The model was fed with clips of $T=50$ seconds, using an embedding dimension of $d=512$. We utilized $|\mathcal{E}|=6$ embeddings, 5 corresponding to visual data, and the remaining one to audio data. The temporal dimension of the input visual features was set to $L_{j} = T, \hspace{1mm} \forall j \in \{1, ..., |\mathcal{E}|-1\}$, while for the audio $L_{a} = \lfloor T / T_{a}\rfloor$ with $T_{a} = 0.96$ seconds. Furthermore, we set the temporal output dimension as $L_{out}=2 \times T$. In our model, we employed $n_{e}=3$, $n_{d}=3$, $h=8$ and $\alpha=4$ along with $C=17$. Additionally, we set $p=0.4$ for dropout. The radii were set experimentally as $r_{c}=2$ and $r_{d}=3$ seconds, and the loss function parameters were set as $\gamma = 1$, $\alpha_{L} = 0.3$, and $w_{c}=100$. We also applied balanced mixup with parameters $\alpha_{m}=1$ and $\beta_{m}=0.6$, along with $p_{td} = 0.5$ and $p_{ts} = 0.3$. The window size for Soft Non-Maximum Suppression was experimentally determined for each action, with values ranging from 5 to 14 seconds. Code is available at \url{https://github.com/arturxe2/ASTRA}.

\subsection{Ablations}
\label{sec:ablations}

The results in terms of tight and loose Average-mAP are summarized in Table~\ref{tab:results_test}. These results are obtained on the Test split following the evaluation protocol P1.

\begin{table}[ht]
	\centering
	\scalebox{0.81}{
		\begin{tabular}{|llcc|}\hline
			\emph{Model} & \emph{Added feature} & \emph{loose A-mAP} & \emph{tight A-mAP}\\
			\hline\hline
			\multicolumn{4}{|l|}{\textbf{Base models}}\\
			M0 & - & 75.21 & 62.38 \\
            M1 & M0 + Hierarchical TE &  75.42 & 62.32 \\
			M2 & M1 + $r_{c}=2, r_{d}=3$ & 74.65 & 63.97 \\
			\hline
            \multicolumn{4}{|l|}{\textbf{Data Augmentations}}\\
			M3 & M2 + Mixup (0.6, 0.6) & 75.61 & 64.97 \\
			M4 & M2 + Balanced mixup (1, 0.6) & 76.55 & 65.82 \\
            M5 & M4 + Other augmentations & 77.49 & 66.07 \\
		    \hline
		    \multicolumn{4}{|l|}{\textbf{Output dimension}}\\
			M6 & M5 + $L_{out}=T$ & 77.72 & 64.24\\
            \hline
            \multicolumn{4}{|l|}{\textbf{Additional improvements}}\\
            M7 & M5 + Focal loss & 78.02 & 66.09 \\
            M8 & M7 + Uncertainty & 78.14 & 66.63 \\
			\textbf{M9 (ASTRA)} & M8 + Audio modality & \textbf{78.09} & \textbf{66.82} \\
			\hline
		\end{tabular}
	}
	\caption{Ablation experiments with Average-mAP results on the Test split under P1 evaluation.}
	\label{tab:results_test} 
	\vspace{-0.5cm}
\end{table}

The base model (M0) is the simplest and differs significantly from our solution ASTRA. It only utilizes visual embeddings and replaces the Hierarchical Transformer encoder with a vanilla Transformer encoder. Moreover, it does not employ any data augmentation techniques. Additionally, it lacks the focal loss term in the classification loss, and the uncertainty-aware displacement head is substituted with a typical regression head that utilizes mean squared error as the displacement loss. Furthermore, it utilizes the optimal values in Soares et al.~\cite{yahoo} for the radii, $r_{c}=3$ and $r_{d}=6$. This model achieves a tight Average-mAP of 62.38.

In M1, we introduce the Hierarchical Transformer encoder. As shown in Table~\ref{tab:results_test}, the results are comparable to M0. There is a slight increase in performance on the loose Average-mAP by +0.21, while a minor reduction is observed in the tight metric by -0.06. Despite these similar results, we opt to use the Hierarchical Transformer encoder as it offers computational cost reduction compared to the vanilla Transformer encoder. Moreover, through experimental modifications, we adjust the radii of action detection to $r_{c}=2$ and $r_{d}=3$ resulting in an improvement of +1.65 on the tight metric. 

\begin{figure}[ht]
	\centering
	\includegraphics[width=1\linewidth]{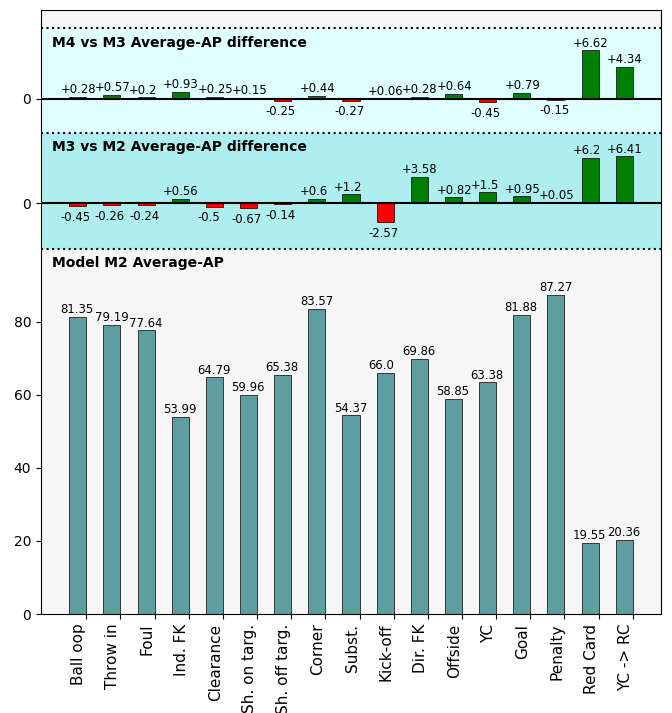}
	%   \caption{Base model for action spotting. Multimodal transformer with audio and video features and intermediate supervision.}
	\caption{Per-class results comparison of models M2, M3, and M5. The figure displays Average-AP scores for each action in M2 (bottom), the differences between M2 and M3 (middle), and the differences between M3 and M4 (top).}
    \label{fig:m2vsm3vsm4}
    \vspace{-0.6cm}
\end{figure}

As anticipated, the introduction of data augmentation techniques, such as typical mixup, enhances the model's generalization capabilities. Specifically, when using the best set of tried parameters $(\alpha_{m}, \beta_{m}) = (0.6, 0.6)$ in M3, we observe an additional improvement of +1.00 in the tight Average-mAP. Figure~\ref{fig:m2vsm3vsm4} illustrates that these improvements are most pronounced in tail actions, such as red cards or second yellow cards. This can be attributed to the fact that tail actions are more prone to overfitting, thus benefiting greatly from improved generalization. By adapting the typical mixup approach to our proposed balanced mixup, and utilizing the best of the tried parameter combinations $(\alpha_{m}, \beta_{m}) = (1, 0.6)$, we achieve an additional improvement of +0.85. As depicted in Figure~\ref{fig:m2vsm3vsm4}, these improvements are observed across most of the action classes, although the changes are relatively smaller in head classes. Once again, the impact on tail classes is particularly notable. These results demonstrate the effectiveness of our proposed balanced mixup technique in handling long-tail data. Furthermore, the introduction of additional augmentations such as temporal dropout and temporal switch leads to a further performance boost of +0.25.

Model M6 serves as a demonstration of the importance of an adequate output temporal dimension. As seen in Table \ref{tab:results_test}, when employing a smaller output temporal dimension (i.e. $T_{out}=T$) there is a noticeable decrease in performance with respect to M5 by -1.84. This finding empowers the use of the Transformer encoder-decoder module that allows the output dimension to not be restricted by the input dimension. Other modifications to M5, such as introducing a focal loss term in the classification loss (M7), also led to a slight improvement in performance, particularly in the loose metric. 

\begin{figure}[ht]
	\centering
	\includegraphics[width=1\linewidth]{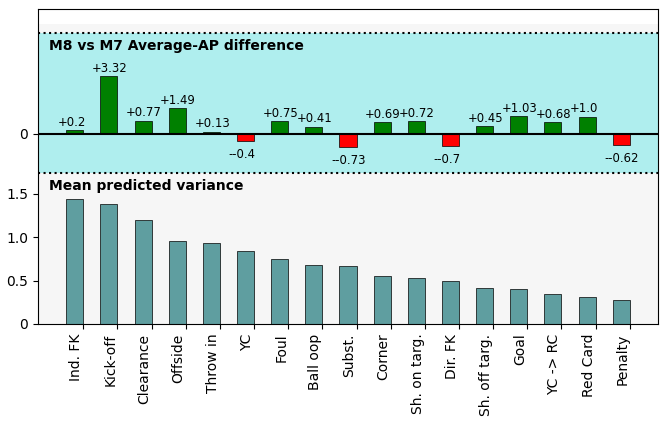}
	%   \caption{Base model for action spotting. Multimodal transformer with audio and video features and intermediate supervision.}
        \vspace{-0.8cm}
	\caption{Analysis of M8 model: mean predicted displacement variance in temporal locations with classification probability greater than 0.5 (bottom) and difference in Average-AP between M7 and M8 (top).}
	\label{fig:variability}
 \vspace{-0.5cm}
\end{figure}

Furthermore, the inclusion of the uncertainty-aware displacement head in M8 resulted in a notable enhancement of +0.54, demonstrating the effectiveness of this module. Figure~\ref{fig:variability} presents a visualization of the average predicted variability associated with each action prediction. Notably, actions with higher variability are primarily those with high non-visibility (e.g., kick-off, clearance, indirect free-kick, throw-in) or actions that require the annotator's judgment for precise temporal annotation, such as offsides. It is in these actions with high variability that the module seems to show the most improvement, supporting our hypothesis that the uncertainty-aware displacement head performs better in handling noisy labels compared to a typical regression head.

\begin{figure}[ht]
	\centering
	\includegraphics[width=1\linewidth]{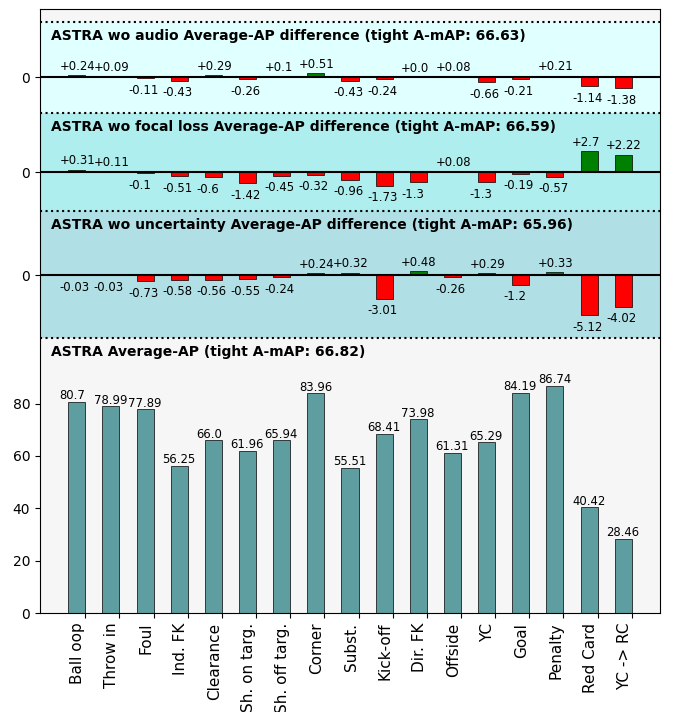}
	%   \caption{Base model for action spotting. Multimodal transformer with audio and video features and intermediate supervision.}
	\caption{Per-class results of ASTRA and ensemble models. The figure illustrates Average-AP scores for each action in ASTRA (bottom), the differences when removing uncertainty (middle-down), differences when removing focal loss (middle-top), and differences when removing audio (top).}
	\label{fig:ensemble}
    \vspace{-0.4cm}
\end{figure}

Finally, the inclusion of audio in M9 further enhances the model, contributing an additional +0.19 improvement and resulting in a 66.82 tight Average-mAP for the ASTRA model. In Figure~\ref{fig:ensemble} (bottom), we can observe the diverse scores for each individual action. \\

\noindent {\large\textbf{Ensemble of ASTRAs.}} To further enhance the results for the SoccerNet Action Spotting Challenge 2023, we explore the use of an ensemble comprising modifications of ASTRA models. As shown in Figure~\ref{fig:ensemble}, the removal of different aspects of ASTRA leads to models that maintain strong overall performance while exhibiting diverse predictions. Each of the models demonstrates improved performance for specific actions. The diversity among the models within the ensemble is crucial for achieving effective ensembling. With this in mind, we propose an ensemble that combines our final ASTRA model with the models depicted in Figure~\ref{fig:ensemble}. These additional models remove audio, focal loss, and uncertainty components, respectively. For each temporal position, we average the predictions of all models in the ensemble. By employing this ensemble approach, we achieve a tight Average-mAP of 67.60 (+0.78). This result emphasizes the ability of appropriately diverse models in an ensemble to provide a slight improvement over the individual performance of ASTRA.

%\vspace{0.4cm}
\subsection{Results on challenge split}

For the evaluation of ASTRA models in the challenge split, we followed the evaluation protocol P2. The results, presented in Table \ref{tab:results_chall}, showcase ASTRA's performance in comparison to the top models in the SoccerNet 2023 Action Spotting challenge. Notably, ASTRA achieves a tight Average-mAP score of 69.43. With the implementation of the ensemble approach, we observe a further improvement, reaching an Average-mAP of 70.21. This result secures the 3rd position in the challenge, surpassing the previous baseline by a +1.88 margin. It is worth noting that ASTRA's performance stands close to the winning solutions, with a difference gap of 1.10 points from the current SOTA. Additionally, our method achieves the best results on the loose metric and on non-visible actions. The incorporation of label uncertainty modeling and the inclusion of audio input likely contribute to these results, especially in scenarios where label noise is pronounced for non-visible actions. 

\begin{table}[ht]
	\centering
	\scalebox{0.78}{
		\begin{tabular}{|l|ccc|ccc|}
        \hline
        \multirow{2}{*}{\emph{Model}} & \multicolumn{3}{c|}{Tight Average-mAP} & \multicolumn{3}{c|}{Loose Average-mAP} \\ \cline{2-7} 
         & \multicolumn{1}{c|}{\emph{All}} & \emph{Vis.} & \emph{Non vis.} & \multicolumn{1}{c|}{\emph{All}} & \emph{Vis.} & \emph{Non vis.} \\ \hline
        1- SDU\_VSISLAB & \multicolumn{1}{c|}{\textbf{71.31}} & 76.29 & 54.09 & \multicolumn{1}{c|}{78.56} & 81.67 & 69.13 \\
        2- mt\_player & \multicolumn{1}{c|}{71.10} & \textbf{77.22} & 58.5 & \multicolumn{1}{c|}{78.79} & \textbf{82.02} & 77.62 \\ \hline
        3a- \textbf{ASTRA (ensemble)}* & \multicolumn{1}{c|}{70.21} & 75.08 & \textbf{62.34} & \multicolumn{1}{c|}{\textbf{79.27}} & 81.85 & 79.39 \\
        3b- \textbf{ASTRA} & \multicolumn{1}{c|}{69.43} & 74.40 & 61.10 & \multicolumn{1}{c|}{79.02} & 81.70 & \textbf{79.47} \\ \hline
        4- team\_ws\_action & \multicolumn{1}{c|}{69.17} & 75.18 & 59.12 & \multicolumn{1}{c|}{76.95} & 80.39 & 75.92 \\
        5- CEA\_LVA & \multicolumn{1}{c|}{68.38} & 74.79 & 47.68 & \multicolumn{1}{c|}{73.98} & 78.57 & 61.75 \\
        Baseline- Yahoo~\cite{soares2022action} & \multicolumn{1}{c|}{68.33} & 73.22 & 60.88 & \multicolumn{1}{c|}{78.06} & 80.58 & 78.32 \\ \hline
		\end{tabular}
	}
	\caption{Comparison of ASTRA with other state-of-the-art models in terms of Average-mAP in the SoccerNet 2023 challenge. Metrics include loose and tight evaluations for all actions (All), visible actions (Vis.), and non-visible actions (Non vis.). Results marked with an asterisk (*) indicate slight differences from the official SoccerNet challenge results due to minor modifications in the ASTRA architecture.}
	\label{tab:results_chall} 
	\vspace{-1cm}
\end{table}

\section{Conclusion}

This work presented ASTRA, a model designed to address the task of action spotting in soccer matches. Ablation studies demonstrate the effectiveness of different modules within the model in tackling the challenges inherent to the task and the dataset, such as the need for precise spots, the long-tail distribution of the data, the non-visibility in some actions, and the issue of noisy labels. Additionally, ASTRA achieves good results in the SoccerNet 2023 Action Spotting challenge. It surpasses the previous SOTA performance by +1.88, and its performance is in close proximity to that of the challenge winners. \\

\noindent {\large\textbf{Acknowledgements.}} This work has been partially supported by the Spanish project PID2022-136436NB-I00 and by ICREA under the ICREA Academia programme.

\newpage
%%
%% The next two lines define the bibliography style to be used, and
%% the bibliography file.
\bibliographystyle{ACM-Reference-Format}
\bibliography{main}

\end{document}